\pdfoutput=1

\documentclass[11pt,a4paper]{article}
\usepackage[hyperref]{acl2020}
\usepackage{times}
\usepackage{url}            
\usepackage{latexsym}

\usepackage{enumitem}

\usepackage{microtype}

\aclfinalcopy 

\title{Towards Faithfully Interpretable NLP Systems: \\ How should we define and evaluate faithfulness?}

\author{Alon Jacovi \\
  Bar Ilan University \\
  \texttt{alonjacovi@gmail.com} \\\And
  Yoav Goldberg \\
  Bar Ilan University and 
  Allen Institute for AI \\
  \texttt{yoav.goldberg@gmail.com}
  }

\date{}

\begin{document}
\maketitle
\begin{abstract}


With the growing popularity of deep-learning based NLP models, comes a need for interpretable systems. But what is interpretability, and what constitutes a high-quality interpretation? In this opinion piece we reflect on the current state of interpretability evaluation research. We call for more clearly differentiating between different desired criteria an interpretation should satisfy, and focus on the faithfulness criteria. We survey the literature with respect to faithfulness evaluation, and arrange the current approaches around three assumptions, providing an explicit form to how faithfulness is ``defined'' by the community. We provide concrete guidelines on how evaluation of interpretation methods should and should not be conducted. Finally, we claim that the current binary definition for faithfulness sets a potentially unrealistic bar for being considered faithful. We call for discarding the binary notion of faithfulness in favor of a more graded one, which we believe will be of greater practical utility.

\end{abstract}

\section{Introduction}

Fueled by recent advances in deep-learning and language processing, NLP systems are increasingly being used for prediction and decision-making in many fields \cite{DBLP:journals/corr/abs-1906-04284}, including sensitive ones such as health, commerce and law \cite{Fort2016YesWC}. Unfortunately, these highly flexible and highly effective neural models are also opaque. 
There is therefore a critical need for explaining learning-based models' decisions.

The emerging research topic of 
 \textit{interpretability} or \textit{explainability}\footnote{Despite fine-grained distinctions between the terms, within the scope of this work we use the terms ``interpretability'' and ``explainability'' interchangeably.} has grown rapidly in recent years. Unfortunately, not without growing pains.



One such pain is the challenge of defining---and evaluating---what constitutes a quality interpretation.
Current approaches define interpretation in a rather ad-hoc manner, motivated by practical use-cases and applications.
However, this view often fails to distinguish between distinct aspects of the interpretation's quality, such as readability, plausibility and faithfulness \cite{DBLP:journals/corr/abs-1711-07414}.\footnote{Unfortunately, the terms in the literature are not yet standardized, and vary widely. ``Readability'' and ``plausibility'' are also referred to as ``human-interpretability'' and ``persuasiveness'', respectively (e.g., \citet{DBLP:journals/corr/abs-1902-00006,DBLP:journals/corr/abs-1711-07414}). To our knowledge, the term ``faithful interpretability'' was coined in \citet{harrington1985harvey}, reinforced by \citet{Ribeiro:2016:WIT:2939672.2939778}, and is, we believe, most commonly used (e.g., \citet{gilpin2018explaining-faithfulness,wu2018faithful,lakkaraju2019faithful}). \citet{accountability} refers to this issue (more or less) as ``accountability''. Sometimes referred to as how ``trustworthy'' \cite{camburu2019i} or ``descriptive'' \cite{Carmona2015TowardsEF,dalex} the interpretation is, or as ``descriptive accuracy'' \cite{Murdoch2019InterpretableML}. Also related to the ``transparency'' \citep{Baan2019DoTA}, the ``fidelity'' \cite{fidelity}  or the ``robustness'' \citep{alvarez2018robustness} of the interpretation method. And frequently, simply ``explainability'' is inferred to require faithfulness by default.}
We argue (\S\ref{faithfulness-vs-plausibility}, \S\ref{guidelines}) such conflation is harmful, and that faithfulness should be defined and evaluated \textit{explicitly}, and independently from plausibility.

Our main focus is the \emph{evaluation of the faithfulness} of an explanation. Intuitively, a faithful interpretation is one that accurately represents the reasoning process behind the model's prediction. We find this to be a pressing issue in explainability: in cases  where an explanation is required to be faithful, imperfect or misleading evaluation can have disastrous effects.

While literature in this area may implicitly or explicitly evaluate faithfulness for specific explanation techniques, there is no consistent and formal definition  of faithfulness. 
We uncover three assumptions that underlie all these attempts. By making the assumptions explicit and organizing the literature around them, we ``connect the dots'' between seemingly distinct evaluation methods, and also provide a basis for discussion regarding the desirable properties of faithfulness (\S\ref{defining-faithfulness}).

Finally, we observe a trend by which faithfulness is treated as a binary property, followed by showing that an interpretation method is not faithful. We claim that this is unproductive (\S\ref{evaluating-faithfulness}), as the assumptions are nearly impossible to satisfy fully, and it is all too easy to disprove the faithfulness of an interpretation method via a counter-example. 
What can be done? We argue for a more practical view of faithfulness, calling for a \emph{graded criteria} that measures \emph{the extent and likelihood} of an interpretation to be faithful, \emph{in practice} (\S\ref{towards}).
While we started to work in this area, we pose the exact formalization of these criteria, and concrete evaluations methods for them, as a central challenge to the community for the coming future. 

\section{Faithfulness vs. Plausibility}
\label{faithfulness-vs-plausibility}

There is considerable research effort in attempting to define and categorize the desiderata of a learned system's interpretation, most of which revolves around specific use-cases
 \cite[inter alia]{lipton2016mythos,fidelity}.
 
Two particularly notable criteria, each useful for a different purposes, are \emph{plausibility} and \emph{faithfulness}.
``Plausibility'' refers to how convincing the interpretation is to humans, 
while ``faithfulness'' refers to how accurately it reflects the true reasoning process of the model \cite{DBLP:journals/corr/abs-1711-07414,wiegreffe2019attentionisnotnot}.

Naturally, it is possible to satisfy one of these properties without the other. For example, consider the case of interpretation via post-hoc text generation---where an additional ``generator'' component outputs a textual explanation of the model's decision, and the generator is learned with supervision of textual explanations \cite{zaidan-eisner-2008-modeling,DBLP:journals/corr/abs-1906-02361,DBLP:journals/corr/abs-1905-13714}. In this case, plausibility is the dominating property, while there is no faithfulness guarantee.

Despite the difference between the two criteria, many authors do not clearly make the distinction, and sometimes conflate the two.\footnote{E.g., \citet{Lundberg2017AUA,DBLP:journals/corr/abs-1801-06422,wu2018faithful}.}
Moreoever, the majority of works do not explicitly name the criteria under consideration, even when they clearly belong to one camp or the other.\footnote{ E.g., \citet{mohseni2018human,DBLP:journals/corr/ArrasHMMS16a,DBLP:journals/corr/abs-1811-03970,DBLP:journals/corr/abs-1907-03324}.} 

We argue that this conflation is dangerous.
For example, consider the case of \emph{recidivism prediction}, where a judge is exposed to a model's prediction and its interpretation, and the judge believes the interpretation to reflect the model's reasoning process. Since the interpretation's faithfulness carries legal consequences, a \emph{plausible} but \emph{unfaithful} interpretation may be the worst-case scenario. 
The lack of explicit claims by research may cause misinformation to potential users of the technology, who are not versed in its inner workings.\footnote{As \citet{Kaur2019InterpretingIU} concretely show, even experts are prone to overly trust the faithfulness of explanations, despite no guarantee.} Therefore, clear distinction between these terms is critical. 


\section{Inherently Interpretable?}

A distinction is often made between two methods of achieving interpretability: (1) \emph{interpreting existing models via post-hoc techniques}; and (2) \emph{designing inherently interpretable models.}
\citet{rudin2018please} argues in favor of \emph{inherently interpretable models}, which by design claim to provide more faithful interpretations than post-hoc interpretation of black-box models.

We warn against taking this argumentation at face-value: a method being ``inherently interpretable'' is merely a claim that needs to be verified before it can be trusted. Indeed, while \emph{attention mechanisms} have been considered as ``inherently interpretable'' \cite{DBLP:journals/corr/abs-1808-03894,lee-etal-2017-interactive}, recent work cast doubt regarding their faithfulness
\cite{sofia-isattentioninterpretable,jain2019attentionisnot,wiegreffe2019attentionisnotnot}. 


\section{Evaluation via Utility}

While explanations have many different use-cases, such as model debugging, lawful guarantees or health-critical guarantees, one other possible use-case with particularly prominent evaluation literature is Intelligent User Interfaces (IUI), via Human-Computer Interaction (HCI), of automatic models assisting human decision-makers. In this case, the goal of the explanation is to increase the degree of trust between the user and the system, giving the user more nuance towards whether the system's decision is likely correct, or not. In the general case, the final evaluation metric is the performance of the user at their task \cite{DBLP:conf/chi/AbdulVWLK18}. For example, \citet{10.1145/3301275.3302265} evaluate various explanations of a model in a setting of trivia question answering. 

However, in the context of faithfulness, we must warn against HCI-inspired evaluation, as well: \textbf{increased performance in this setting is not indicative of faithfulness; rather, it is indicative of correlation between the plausibility of the explanations and the model's performance.}

To illustrate, consider the following fictional case of a non-faithful explanation system, in an HCI evaluation setting: the explanation given is a heat-map of the textual input, attributing scores to various tokens. Assume the system explanations behave in the following way: when the output is \textit{correct}, the explanation consists of random content words; and when the output is \textit{incorrect}, it consists of random punctuation marks. In other words, the explanation is more likely to appear plausible when the model is correct, while at the same time not reflecting the true decision process of the model. The user, convinced by the nicer-looking explanations, performs better using this system. However, the explanation consistently claimed random tokens to be highly relevant to the model's reasoning process. While the system is concretely useful, the claims given by the explanation do not reflect the model's decisions whatsoever (by design). 

While the above scenario is extreme, this misunderstanding is not entirely unlikely, since \textit{any} degree of correlation between plausibility and model performance will result in increased user performance, regardless of any notion of faithfulness.

\section{Guidelines for Evaluating Faithfulness}
\label{guidelines}

We propose the following guidelines for evaluating the faithfulness of explanations. These guidelines address common pitfalls and sub-optimal practices we observed in the literature.

\paragraph{Be explicit in what you evaluate.}
Conflating plausability and faithfulness is harmful. You should be explicit on which one of them you evaluate, and use suitable methodologies for each one. Of course, the same applies when designing interpretation techniques---be clear about which properties are being prioritized.


\paragraph{Faithfulness evaluation should not involve human-judgement on the quality of interpretation.}
We note that: (1) humans cannot judge if an interpretation is faithful or not: if they understood the model, interpretation would be unnecessary; 
(2) for similar reasons, we cannot obtain supervision for this problem, either. Therefore, human judgement should not be involved in evaluation for faithfulness, as human judgement measures plausability.

\paragraph{Faithfulness evaluation should not involve human-provided gold labels.}
We should be able to interpret incorrect model predictions, just the same as correct ones. Evaluation methods that rely on gold labels are influenced by human priors on what \emph{should} the model do, and again push the evaluation in the direction of plausability.

\paragraph{Do not trust ``inherent interpretability'' claims.}
Inherent interpretability is a claim until proven otherwise.
Explanations provided by ``inherently interpretable'' models must be held to the same standards as post-hoc interpretation methods, and be evaluated for faithfulness using the same set of evaluation techniques.

\paragraph{Faithfulness evaluation of IUI systems should not rely on user performance.} End-task user performance in HCI settings is merely indicative of correlation between plausibility and model performance, however small this correlation is. While important to evaluate the utility of the interpretations for some use-cases, it is unrelated to faithfulness.

\section{Defining Faithfulness} \label{defining-faithfulness}

What does it mean for an interpretation method to be faithful? Intuitively, we would like the provided interpretation to reflect the true reasoning process of the model when making a decision. But what is a reasoning process of a model, and how can reasoning processes be compared to each other?

Lacking a standard definition, different works evaluate their methods by introducing tests to measure properties that they believe good interpretations should satisfy. Some of these tests measure aspects of faithfulness. These ad-hoc definitions are often unique to each paper and inconsistent with each other, making it hard to find commonalities.

We uncover \emph{three assumptions} that underlie all these methods, enabling us to organize the literature along standardized axes, and relate seemingly distinct lines of work. Moreover, exposing the underlying assumptions enables an informed discussion regarding their validity and merit (we leave such a discussion for future work, by us or others).

These assumptions, to our knowledge, encapsulate the current working definitions of faithfulness used by the research community.

\paragraph{Assumption 1 (\textit{The Model Assumption}).}{
\emph{Two models will make the same predictions if and only if they use the same reasoning process.}
}\\[0.3em] 
\noindent\textbf{Corollary 1.1.} \ \ \ \textit{An interpretation system is unfaithful if it results in different interpretations of models that make the same decisions.}

As demonstrated by a recent example concerning NLP models, it can be used for proof by counter-example. Theoretically, if all possible models which can perfectly mimic the model's decisions also provide the same interpretations, then they could be deemed faithful. Conversely, showing that two models provide the same results but different interpretations, disprove the faithfulness of the method. \citet{wiegreffe2019attentionisnotnot} show how these counter-examples can be derived with adversarial training of models which can mimic the original model, yet provide different explanations.\footnote{We note that in context, \citeauthor{wiegreffe2019attentionisnotnot} also utilize the model assumption to show that some explanations do carry useful information on the model's behavior.}
\\[0.3em] 
\noindent\textbf{Corollary 1.2.} \ \ \ \textit{An interpretation is unfaithful if it results in different decisions than the model it interprets.}

A more direct application of the \textit{Model Assumption} is via the notion of \emph{fidelity} \cite{fidelity,lakkaraju2019faithful}. For cases in which the explanation is itself a model capable of making decisions (e.g., decision trees or rule lists \cite{sushil-etal-2018-rule}), \emph{fidelity} is defined as the degree to which the explanation model can mimic the original model's decisions (as an accuracy score). For cases where the explanation is \emph{not} a computable model, \citet{doshi2017rigorous-humangrounded} propose a simple way of mapping explanations to decisions via crowd-sourcing, by asking humans to simulate the model's decision without any access to the model, and only access to the input and explanation (termed \textit{forward simulation}). This idea is further explored and used in practice by \citet{nguyen-2018-comparing}.

\paragraph{Assumption 2 (\textit{The Prediction Assumption}).}{
\textit{On similar inputs, the model makes similar decisions if and only if its reasoning is similar.}
}\\[0.3em] 
\noindent\textbf{Corollary 2.} \ \ \ \textit{An interpretation system is unfaithful if it provides different interpretations for similar inputs and outputs.}

Since the interpretation serves as a proxy for the model's ``reasoning'', it should satisfy the same constraints. In other words, interpretations of similar decisions should be similar, and interpretations of dissimilar decisions should be dissimilar. 

This assumption is more useful to \emph{disprove} the faithfulness of an interpretation rather than prove it, since a disproof requires finding appropriate cases where the assumption doesn't hold, where a proof would require checking a (very large) satisfactory quantity of examples, or even the entire input space.

One recent discussion in the NLP community \cite{jain2019attentionisnot,wiegreffe2019attentionisnotnot} concerns the use of this underlying assumption for evaluating attention heat-maps as explanations. The former attempts to provide different explanations of similar decisions \emph{per instance}. The latter critiques the former and is based more heavily on the \emph{model assumption}, described above. 


Additionally, \citet{kindermans2017unreliability-of-saliency} propose to introduce a constant shift to the input space, and evaluate whether the explanation changes significantly as the final decision stays the same. \citet{alvarez2018robustness} formalize a generalization of this technique under the term \textit{interpretability robustness}: interpretations should be invariant to small perturbations in the input (a direct consequence of the \emph{prediction assumption}). \citet{DBLP:conf/aies/WolfGH19} further expand on this notion as ``consistency of the explanation with respect to the model''.  Unfortunately, robustness measures are difficult to apply in NLP settings due to the discrete input. 



\paragraph{Assumption 3 (\textit{The Linearity Assumption}).\footnote{This assumption has gone through justified scrutiny in recent work. As mentioned previously, we do not necessarily endorse it. Nevertheless, it is used in parts of the literature.}}
\emph{Certain parts of the input are more important to the model reasoning than others. Moreover, the contributions of different parts of the input are independent from each other.}
\\[0.3em] 
\noindent\textbf{Corollary 3.} \ \ \  \textit{Under certain circumstances, heat-map interpretations can be faithful.}

This assumption is employed by methods that consider heat-maps\footnote{Also referred to as feature-attribution explanations \cite{kim2017interpretability}.} (e.g., attention maps) over the input as explanations,
particularly popular in NLP.
Heat-maps are \emph{claims} about which parts of the input are more relevant than others to the model's decision.
As such, we can design ``stress tests'' to verify whether they uphold their claims.


One method proposed to do so is \textit{erasure}, where the ``most relevant'' parts of the input---according to the explanation---are erased from the input, in expectation that the model's decision will change \citep{DBLP:journals/corr/ArrasHMMS16a,feng2018pathologies,sofia-isattentioninterpretable}. Otherwise, the ``least relevant'' parts of the input may be erased, in expectation that the model's decision will not change \citep{jacovi2018understanding}.
\citet{DBLP:journals/corr/abs-1910-13294,eraser2019} propose two measures of \emph{comprehensiveness} and \emph{sufficiency} as a formal generalization of erasure: as the degree by which the model is influenced by the removal of the high-ranking features, or by inclusion of solely the high-ranking features. 



\section{Is Faithful Interpretation Impossible?}
\label{evaluating-faithfulness}

The aforementioned assumptions are currently utilized to evaluate faithfulness in a binary manner, whether an interpretation is strictly faithful or not. 
Specifically, they are most often used to show that a method is \emph{not} faithful, by constructing cases in which the assumptions do not hold for the suggested method.\footnote{Whether for attention \cite{Baan2019DoTA,Pruthi2019LearningTD,jain2019attentionisnot,sofia-isattentioninterpretable,wiegreffe2019attentionisnotnot}, saliency methods \cite{alvarez2018robustness,kindermans2017unreliability-of-saliency}, or others \cite{ghorbani2019interpretation,feng2018pathologies}.}
In other words, \textbf{there is a clear trend of proof via counter-example, for various interpretation methods, that they are not globally faithful.} 

We claim that this is unproductive, as we expect these various methods to consistently result in negative (not faithful) results, continuing the current trend. 
This follows because an interpretation functions as an \emph{approximation} of the model or decision's true reasoning process, so it by definition loses information. 
By the pigeonhole principle, there will be inputs with deviation between interpretation and reasoning. 

This is observed in practice, in numerous work that show adversarial behavior, or pathological behaviours, that arise from the 
deeply non-linear and high-dimensional decision boundaries of current models.
\footnote{\citet[\S 6]{kim2017interpretability,feng2018pathologies} discuss this point in the context of heat-map explanations.}
Furthermore, because we lack supervision regarding which models or decisions are indeed mappable to human-readable concepts, we cannot ignore the approximation errors.


This poses a high bar for explanation methods to fulfill, a bar which we estimate will not be overcome soon, if at all. What should we do, then, if we desire a system that provides faithful explanations?


\section{Towards Better Faithfulness Criteria} \label{towards}
We argue that a way out of this standstill is in a more practical and nuanced methodology for defining and evaluating faithfulness.
We propose the following challenge to the community:
\textbf{We must develop formal definition and evaluation for faithfulness that allows us the freedom to say when a method is \emph{sufficiently faithful} to be useful in practice.}

We note two possible approaches to this end: 

\begin{enumerate}[itemsep=0pt]
    \item \textbf{Across models and tasks:} The \emph{degree} (as grayscale) of faithfulness at the level of specific models and tasks. Perhaps some models or tasks allow sufficiently faithful interpretation, even if the same is not true for others.\footnote{As noted by \citet{wiegreffe2019attentionisnotnot,anonymous2020attention}, although in the context of attention solely.} \\
    For example, the method may not be faithful for some question-answering task, but faithful for movie review sentiment, perhaps based on various syntactic and semantic attributes of those tasks.
    \item \textbf{Across input space:} The degree of faithfulness at the level of subspaces of the input space, such as neighborhoods of similar inputs, or singular inputs themselves. If we are able to say with some degree of confidence whether a specific decision's explanation is faithful to the model, even if the interpretation method is not considered universally faithful, it can be used with respect to those specific areas or instances only.
\end{enumerate}


\section{Conclusion}

The opinion proposed in this paper is two-fold:

First, interpretability evaluation often conflates evaluating faithfulness and plausibility together. We should tease apart the two definitions and focus solely on evaluating faithfulness without any supervision or influence of the convincing power of the interpretation. 

Second, faithfulness is often evaluated in a binary ``faithful or not faithful'' manner, and we believe strictly faithful interpretation is a ``unicorn'' which will likely never be found. We should instead evaluate faithfulness on a more nuanced ``grayscale'' that allows interpretations to be useful even if they are not globally and definitively faithful. 


\section*{Acknowledgements}

We thank Yanai Elazar for welcome input on the presentation and organization of the paper. We also thank the reviewers for additional feedback and pointing to relevant literature in HCI and IUI. 

This project has received funding from the Europoean Research Council (ERC) under the Europoean Union's Horizon 2020 research and innovation programme, grant agreement No. 802774 (iEXTRACT).

\bibliography{acl2020}
\bibliographystyle{acl_natbib}

\end{document}